# pLitterStreet: Street Level Plastic Litter Detection and Mapping

Sriram Reddy Mandhati[1*], N. Lakmal Deshapriya[1], Chatura Lavanga Mendis[1], Kavinda Gunasekara[1], Frank Yrle[1], Angsana Chaksan[1], and Sujit Sanjeev[2]

[1]GeoInformatics Center, Asian Institute of Technology, Khlong Nueng, Khlong Luang District, Pathum Thani 12120, Thailand

[2]Independent Researcher

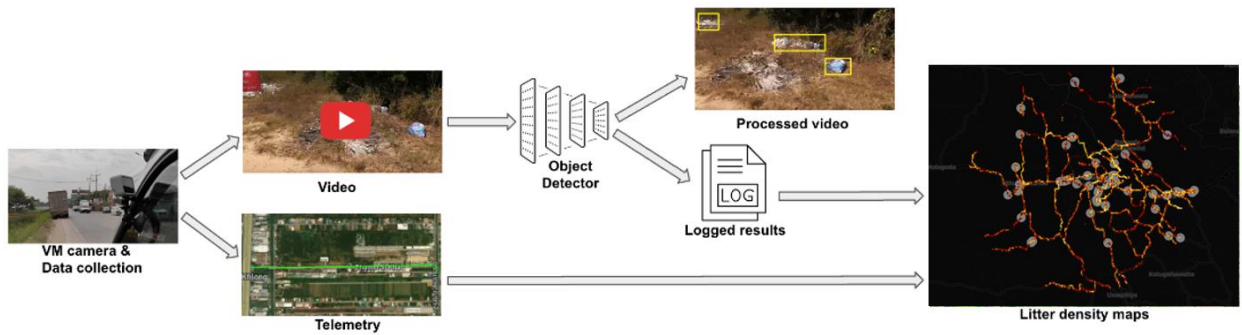

**Fig. 1** Pipeline of plastic litter mapping

**Abstract**

Plastic pollution is a critical environmental issue, and detecting and monitoring plastic litter is crucial to mitigate its impact. This paper presents the methodology of mapping street-level litter, focusing primarily on plastic waste and the location of trash bins. Our methodology involves employing a deep learning technique to identify litter and trash bins from street-level imagery taken by a camera mounted on a vehicle. Subsequently, we utilized heat maps to visually represent the distribution of litter and trash bins throughout cities. Additionally, we provide details about the creation of an open-source dataset ("pLitterStreet") which was developed and utilized in our approach. The dataset contains more than 13,000 fully annotated images collected from vehicle-mounted cameras and includes bounding box labels. To evaluate the effectiveness of our dataset, we tested four well known state-of-the-art object detection algorithms (Faster R-CNN, RetinaNet, YOLOv3, and YOLOv5), achieving an average precision (AP) above 40%. While the results show average metrics, our experiments demonstrated the reliability of using vehicle-mounted cameras for plastic litter mapping. The "pLitterStreet" can also be a valuable resource for researchers and practitioners to develop and further improve existing machine learning models for detecting and mapping plastic litter in an urban environment. The dataset is open-source and more details about the dataset and trained models can be found at https://github.com/gicait/pLitter.

* st120001@alumni.ait.asia

1. Introduction

The longevity of plastic litter in the environment is a major concern due to the release of chemicals during its breakdown, which poses a threat to the surrounding lifeforms. Furthermore, poor plastic litter management practices have led to an increase in littering and illegal dumping, leading to plastic accumulation in drainage systems and flooding of roads and their surroundings. This problem is further compounded by the fact that the littering rate is much higher than the cleanup rate. As such, there is a need for responsible disposal of plastic waste and strict policies to curb plastic pollution.

To address this problem, comprehensive monitoring systems are required to facilitate effective decision-making by local authorities and policy makers. Such monitoring systems can be used to reveal trends in plastic pollution, including affected areas, leakage sources, and illegal dumpsites. This information can act as support towards informed decision-making for plastic waste management and fill the knowledge gaps that exist among plastic producers, consumers, and the authorities.

To enable rapid and effective street-level, city-scale plastic monitoring, we present a standard and deep learning-based methodology to map street-side litter (mainly plastic litter) and trash bins. The data collection process involved the use of a vehicle-mounted camera with simultaneous recording of GPS location. The collected data was processed by separating frames from the recorded videos at regular time intervals. A subset of the resulting data was then annotated by a team of annotators using open-source annotation tools. Subsequently, a deep learning model was trained on annotated data to predict the presence of litter in new video frames. Finally, these predictions along with corresponding GPS locations were utilized to generate heat maps of cities which visualize litter hotspots.

Through this process, we have compiled the pLitterStreet dataset which consists of more than 13,064 images and nearly 78,333 object instances related to litter and 768 trash bin instances. These data were extracted from video feeds captured with vehicle-mounted cameras driven within and at the outskirts of selected cities in Sri Lanka, Thailand, and Vietnam. Notably, the pLitterStreet dataset is the first to utilize street-level imagery or vehicle-mounted cameras for plastic litter detection. Moreover, we included data captured from five cities in three countries, maintaining the diversity of various plastic objects commonly found as litter, including a category for disposable face masks that saw peak usage during the Covid-19 pandemic.

To evaluate the effectiveness of our dataset, we conducted experiments with select popular algorithms and evaluated the resulting accuracy metrics. Our experiments also focused on detecting illegal littering hotspots where piles of litter are in contrast to areas with less concentrated litter. Our results demonstrate the reliability of our approach for plastic litter monitoring using vehicle-mounted cameras and suggest the potential for the application of innovative technologies like computer vision, deep learning, and geospatial data analytics for comprehensive plastic waste monitoring.

2. **Related works**

To our knowledge, the majority of litter mapping studies have primarily focused on crowdsourcing techniques and the creation of litter-related datasets. The following section summarizes previous studies on litter mapping and publicly available litter datasets.

OpenLitterMap[4] is a tool which uses citizen-science approach to map the litter in the street. It provides a mobile application which users can capture and upload litter photos along with geolocation for representation on a map. Optionally, tags for litter in the captured images also can be added before uploading the image. The closest work to this study is a project designed to quantify litter using deep learning techniques from images taken from a city sweeping vehicle mounted camera [5]. However, the dataset is not publicly available.

When it comes to training a deep learning object detection algorithm, it is necessary to prepare a representative dataset with enough diverse examples to achieve a desired level of accuracy – a concept that is congruent with litter detection. Many datasets have been published for the purpose of detecting general objects or items encountered in daily life, for example Microsoft COCO [6], and Pascal VOC [7]. Also a few datasets such as Cityscapes [8], Mapillary Vistas [9],

KITTI [10] presented street-level imagery (or data captured from vehicle-mounted cameras) for the purposes of scene understanding, autonomous driving, or for object detection purposes like pedestrian or traffic light detection.

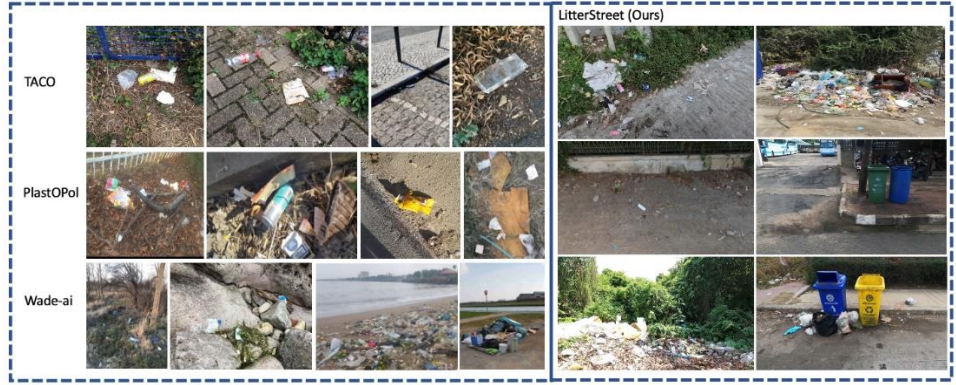

**Fig 2** Comparison of related existing datasets

However, there are fewer studies and datasets such as TACO [11], PlastOPol [12], UAV-BD [13], and UAVVaste [14] published that focus on litter detection with various scenes and background. Some studies such as OpenLitterMap [15] also presented platforms for crowdsourced litter mapping to build datasets and plastic litter evidence. Studies that focused on street level littering and litter found in outdoor settings are explored in the next section.

TACO (Trash Annotations in Context) was prepared with crowdsourced images and annotations of trash in various indoor and outdoor environments. TACO is still growing with continuous contribution through crowdsourcing. The taxonomy of TACO is diverse and broad, covering more than 60 categories of trash types with various scenes or backgrounds. However, not all the classes have enough instances and are reclassified with other classes into broader classes. The size of objects in the images entirely depends on the distance to the camera capturing it, lens specification, and zoom level. Since the dataset is constructed from crowdsourcing, object size varies. Most of the images are captured at close range, lending a high amount of detail to the objects. The images are collected from many background types and are not restricted to street level or outdoor locations.

PlastOPol is another open-source dataset with diverse types of litter samples from various backgrounds including water, sand, snow, flint fields, and streets. All the instances are single class labeled with the name "litter". Based on the bounding box sizes, 90.98 % of the instances are considered large, and only 0.62 % of the instances are small.

The comparison of TACO, PlastOPol, and wade-ai [16] to our dataset is shown in Figure 2. Existing domain-specific datasets are limited in usefulness to our study due to their small size and image orientation different to the problem we are focusing on. Namely, the existing datasets either miss the context of the background or the perspective of the target object. For example, an image of a target litter object at street-side will look different when captured with a handheld camera and a vehicle-mounted camera. The image from the handheld camera is a still photo, with litter likely rendered in focus and highly detailed, especially if captured from a close distance. When the litter object fills the frame, information about the background is lost. Whereas an image captured with a vehicle-mounted camera may not display the litter object with as much detail as the handheld camera but provides sufficient background information. Even if littered locations were captured with sufficient background with handheld cameras, it would require extensive time and labour to collect data at city scale. However, utilizing vehicle-mounted cameras reduces the time and effort spent per image in capturing litter in the streets.

Although these datasets are openly available in the domain of litter/trash/waste/garbage detection, they are: (1) not specific to plastic litter, and (2) not all the images are at street level or roadside. Most of the images are captured from close range with the target object at the center of the frame. As our approach is to leverage the advantages of both the object detection algorithms and vehicle-mounted cameras to speed up city-wide surveying, constructing a suitable dataset that focuses mainly on plastic litter detection at the street level is in order.

3. **Our Approach**

Our methodology is comprised of four main steps. First, data was collected using a vehicle-mounted camera. Second, a subset of the resulting data was annotated using open-source annotation tools. Third, a deep learning model was trained on the annotated data to identify the presence of litter in new video frames. Finally, the predicted litter locations were combined with corresponding GPS locations to produce heat maps of cities, highlighting areas with high litter accumulation. Further details on each of these steps are provided below.

### 3.1. Data Collection

As this study focuses on detecting street-side litter to identify leakage sources into waterways, we needed a feasible and efficient approach to satisfy city-scale data collection. Conditions for data collection included: (1) the ability to capture a large amount of data spanning an area the size of a city to form a large set of examples, (2) being time and cost effective, (3) minimal labor effort to collect data. These conditions led us to implement vehicle-mounted cameras to collect data.

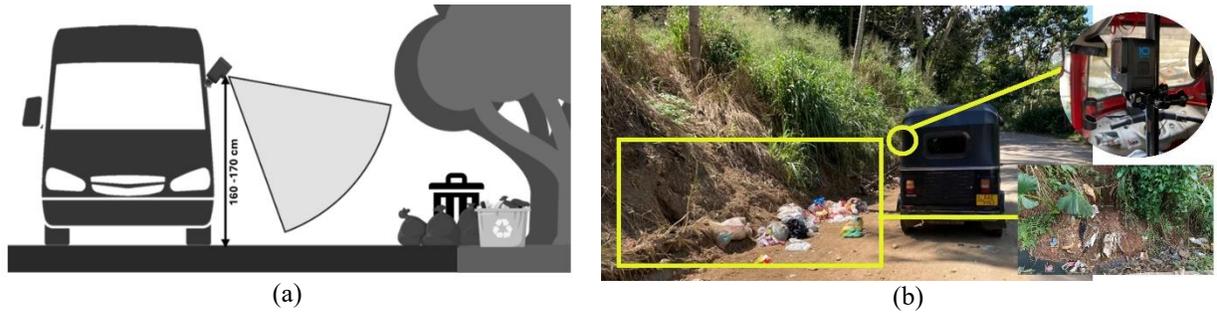

(a)　　　　　　　　　　　　　　　　　　(b)

**Fig 3** Vehicle mounted camera setup (a) Concept diagram (b) Example camera setup used in Sri Lanka

In addition to recording video with the vehicle-mounted camera, it was also necessary to continually record the location of the camera with GPS so that the data could be geo-spatially visualized later in our practical applications. Therefore, a GoPro Hero 9 camera was selected which has an internal GPS module and tags the video feed with geographic coordinates. The compact design of the camera is also a key advantage for easy mounting and handling.

Performing analysis on data collected with vehicle-mounted cameras is a common application of computer vision. Computer vision has widely been used to detect road lanes, traffic lights, pedestrians, and other objects that might be encountered on the road. In the case of litter detection, the perspective of the camera changes depending on the target. As most of the litter is found at the street-side rather than in the middle of the road, the camera should be mounted in a way that the camera lens faces the street-side. An example of the camera orientation in relation to the vehicle and street-side is shown in Figure 3 (a). In Figure 3(b) we show Go Pro Hero 9 camera mounted to a tuk-tuk for data collection in Sri Lanka. Objects in the bottom of the image frame are closer to the vehicle while objects at the top of the images are farther from the vehicle.

The camera height and angle (vertical tilt) were adjusted to center the images to the ground on the street-side. After a few trials, we determined the best camera height to be between 160 to 170 cm. The vertical camera angle was adjusted to bring the ground at street-side into frame, as the driving line is not always the same which causes variation in the distance between the vehicle and street-side. The vehicle's moving speed is also an important factor to consider. Some objects are difficult to identify in the images due to their size or surrounding environment. At excessive speeds, those objects appear blurry and unrecognizable. So, the vehicle speed was kept around 30 kmph except for special situations due to traffic and restrictions. Vehicle-mounted street-side images typical of the different routes can be seen in Figure 4.

Initially, video data was collected by recording the footage with cameras mounting to the vehicles while driving through urban and rural areas. Then, frames were extracted from the videos at a rate of one frame per second. Subsequently, images were screened to eliminate redundancy and extremely blurry images.

### 3.2. Image Annotation

We have used an open-source annotation tool [17] to prepare the labels. The labels are made with bounding boxes and assigned to a category. Image annotation took place over several rounds and was also inspired by how the popular deep learning datasets are annotated.

In the first phase, all of the plastic litter was annotated into one single category as "plastic" since it was not clear about the standard taxonomy of plastic litter. Also, the statistics of litter types in detail, such as plastic bottle, cup, and others were not clear initially. In this trial our main goal was to conduct experiments to prepare a pilot dataset and train a baseline model for checking whether our approach would work, as well as to figure out the possibilities to expand the study.

In the second trial, we added three more categories: face mask, pile, and trash bin. Face masks were added as we observed many face masks littered along streets. The trash bin category was introduced too since it was very helpful in understanding the relationship between trash bin locations and littering locations. In some situations, it was not possible to draw a boundary (annotate an object, bounding box in this case) around all plastic objects if they were found in a cluster. These kinds of instances were separated from other objects by adding a new category called "pile". So, all the clustered objects that fall into this category are drawn into a common bounding box as a single object as a pile.

Additionally, the following guidelines were given to the annotators prior to the beginning of annotation. The annotation guidelines were: (1) the target object must be litter, and it should not be in function nor kept there for a purpose, (2) the bounding box must be strict to the boundaries of the target object location in the images, (3) in the case of large clusters of litter where annotating individual pieces is not feasible, label the whole cluster as a pile with a single bounding box, (4) any box or container that is installed for the purpose of garbage collection can be annotated as "trash bin".

In the third and final trial, we tried to reassign the plastic category into six new labels, "plastic bag", "bottle", "cup", "rope", "sachet", and "straw" which were the plastic types most commonly occurring in the dataset. Still, there were many objects which could not be identified as any of these categories because they were broken or torn plastics or covered in dirt thus reducing capacity for human interpretation. Hence, we kept those objects as a separate "other plastic" category.

Even though the guidelines for annotators were intentionally simple, interpretation of plastic litter can be a challenging task. In some cases, non-plastic items like leaves, paper, and rocks appear very similar to plastic litter. Whether an object is litter or not depends solely on the context rather than a distinct shape or texture. For example, a plastic bottle in one context can be considered litter, while the same plastic object in a different context can be viewed as an ordinary drinking water bottle. This confusion mainly arrives when plastic barrels are used as trash bins and also for other purposes such as storing water. This leads to confusion among annotators, and providing general guidelines for them is challenging. We have shown several example annotations during the annotator training session to better explain what to annotate and what not to annotate. Expert annotators provided annotation verification at the final stage of the annotation process. The final expert review substantially improved data quality, but some interpretation discrepancies were expected in the dataset.

In Figure 4, we can see the warped samples of 9 classes of plastic litter and trash bins from our dataset. Each sample is cropped to its boundaries.

### 3.3. Training

Object detection, a crucial component in the field of computer vision, has seen significant advancements through the application of deep learning algorithms. Over time, numerous supervised algorithms have been developed and proposed, demonstrating reliable improvements in performance. Among these, Fast R-CNN [1], YOLO-v3 [2], and RetinaNet [3] are some of the most widely used. We used Faster R-CNN, RetinaNet, YOLO-v3, and YOLO-v5 [18] to train and evaluate on our dataset. We conducted the experiments on both "4-class" (face mask, pile, plastic, and trash bin) and "10-class" (bag, bottle, cup, face mask, other plastic, pile, rope, sachet, straw, trash bin) versions of the dataset.

All of the experiments were conducted on the Nvidia GeForce GTX 1080 Ti GPU machine. The input image size is scaled to 1024 square pixels for both training and inference. Faster R-CNN and RetinaNet are trained and tested using the implementation by Detectron2 [19] framework. ResNet-101 and ResNet-101-FPN were used as backbones in Faster R-CNN and RetinaNet respectively. PyTorch [21] based implementations were used for training and evaluation of YOLOv3 [20] and YOLOv5 [18]. Anchor boxes used in YOLOv3 and YOLOv5 were generated using the K-Means clustering on the training set samples in our dataset.

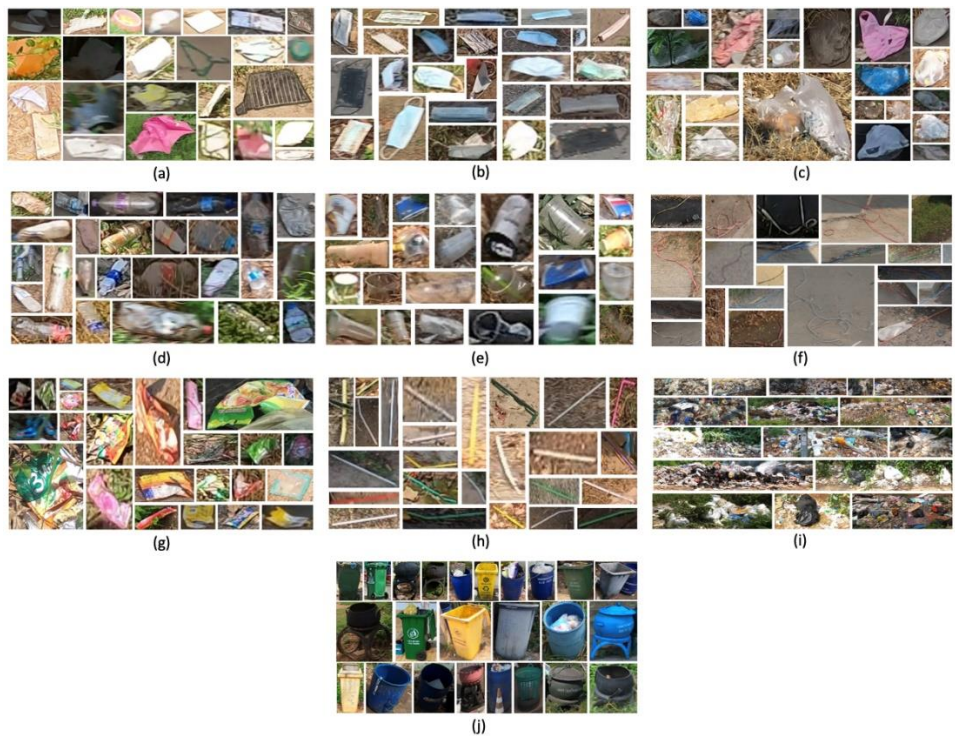

**Fig 4** Warped examples of litters. (a) plastic (miscellaneous), (b) face masks, (c) plastic bag, (d) plastic bottles, (e) plastic cups, (f) ropes, (g) sachets, (h) straws, (i) piles (litter), and (j) trash bins. The examples cropped to bounding boxes from original images and collated together

### 3.4. Heatmap Generation

In the first step, GoPro cameras were used to collect the videos and GPS tracks by mounting them in a vehicle (Cars in our case), and these videos are saved into MP4 format while GPS records are time synchronized and kept in the same file. Frames from videos were extracted with corresponding GPS locations, and a selected object detector model (Faster R-CNN) was used to predict litter and trash bins in each frame. Litter heat maps comprise of two distinct classes: isolated plastic objects and piles. Notably, piles are considerably larger than isolated plastic objects. To ensure visual clarity, we have amplified the representation of piles by multiplying their count in an image by 100, while the count of isolated plastic objects is multiplied by 1. Finally, all detection results and corresponding GPS location data were stored in GeoJSON format to be used in the process of heat map generation.

Since we are interested in the distribution of street-side litter and trash bins, heat maps are a practical way to visualize them. Furthermore, we are interested in the exact location distribution of streetside litter, so color maps were used to visualize streetside litter rather than interpolation-based heatmap generation methods such as Inverse Distance Weighting (IDW) or Kernel Density method. In our case, we have used the "Fire Color Scheme" to visualize street-side litter. Locations with a high amount of streetside litter are visualized in yellow color, while locations with the least amount of street-side litter are visualized in reddish color. On the other hand, trash bin locations are visualized as points large enough to see their distribution with respect to the density of streetside litter. This way, litter hotspots with inadequate trash bins can be easily identified. The overall pipeline to generate the litter density maps is shown in Figure 1.

## 4. Results
### 4.1. Dataset statistics and evaluations

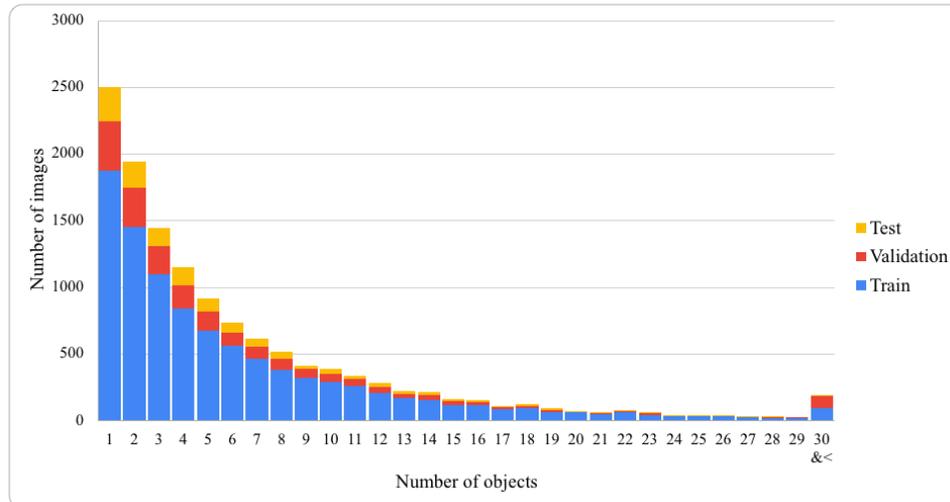

(a)

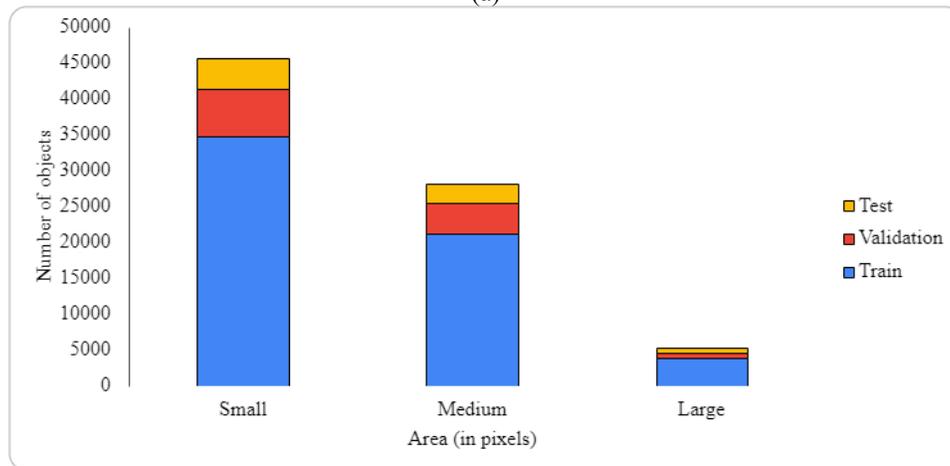

(b)

**Fig 5** pLitterStreet dataset statistics. (a) number of objects per image, and (b) composition of the objects in three different object size ranges

The pLitterStreet dataset continues to receive contributions and refinements to taxonomy. At the time of writing, the dataset used for the experiments consisted of 13,064 images and 79,101 annotations representing 10 classes. The dataset was split into 75, 15, and 10 ratios for train, validation, and test sets. The taxonomy of pLitterStreet is still expanding to fit global plastic litter objects. The present list of categories and their instance count in the pLitterStreet dataset are given in Table 1. The distribution of the number of objects per frame and the distribution of sizes of targets in the datasets are shown in Figure 5(a) and 5(b) respectively. Objects with a size area less than $32^2$ are considered as small, an area in between $32^2$ and $64^2$ is considered as medium, and an area with more than $96^2$ is considered as large. It can be seen that the dataset is dominated by smaller size objects.

**Table 1** Number of targets per category.

| Category      | Train | Validation | Test |
|---------------|-------|------------|------|
| bag           | 6197  | 1216       | 818  |
| bottle        | 1409  | 281        | 206  |
| cup           | 708   | 150        | 77   |
| face mask     | 618   | 136        | 68   |
| pile          | 926   | 209        | 160  |
| rope          | 989   | 166        | 141  |
| sachet        | 2706  | 541        | 367  |
| straw         | 1498  | 264        | 199  |
| trash bin     | 569   | 123        | 76   |
| other plastic | 44151 | 8528       | 5604 |

In our evaluation, we used COCO [6] metrics AP and $AP_{50}$ to measure the accuracy of the models. AP (Average Precision), is calculated by averaging the precision values at 10 different levels of IOU (Intersection over union) thresholds, ranging from 0.50 to 0.95 with a step increase of 0.005. $AP_{50}$ refers to precision at an IOU threshold of 0.50. In multi-class object detections, we calculate the mean values of $AP_{50}$ and AP for each category and used these mean values, denoted as $mAP_{50}$ and mAP respectively, to represent the overall accuracy of the model.

**Table 2** Evaluation on pLitterStreet (4-class) test set.

|              | plastic | pile | face-mask | trash-bin | $mAP_{50}$ | mAP  | mAP (small) | mAP (medium) | mAP (large) |
|--------------|---------|------|-----------|-----------|------------|------|-------------|--------------|-------------|
| Faster R-CNN | 31.9    | 40.3 | 39.0      | 52.5      | 61.2       | 40.9 | 16.7        | 29.5         | 34.7        |
| RetinaNet    | 32.1    | **45.4** | 34.7  | **57.5**  | 63.5       | 42.4 | 17.7        | 29.3         | **43.2**    |
| YOLO-v3l     | 37.1    | 15.6 | **62.5**  | 56.2      | 66.4       | 41.0 | -           | -            | -           |
| YOLO-v5l     | **36.8**| 27.5 | 57.6      | 54.0      | **69.0**   | **44.0** | **25.0** | **34.9**     | 40.5        |

The evaluation results of the models trained on 4-class, and 10-class datasets are shown in Table 2, and 3 respectively. According to the results, categories with small size objects such as cups were the most challenging to detect. From the results, we can observe that mAP of small instances is significantly lower than the medium and large instances. Trash bins were detected with high precision despite having fewer samples than the other classes. That might be due to the constant similarities and less variation for trash bins.

**Table 3** Evaluation on pLitterStreet (10-class) test set.

|        | Faster R-CNN | RetinaNet | YOLO-v3l | YOLO-v5l |
|--------|--------------|-----------|----------|----------|
| bag    | 24.1         | **26.1**  | 25.4     | 23.6     |
| bottle | 20.2         | 18.0      | **23.1** | 20.8     |
| cup | **9.3** | 6.8 | 6.8 | 5.62 |
| facemask | 22.2 | 21.7 | **26.4** | 24.4 |
| other plastic | 24.3 | 23.3 | 25.3 | **26.7** |
| pile | **25.3** | 24.8 | 3.3 | 5.17 |
| rope | 22.1 | 18.2 | 20.8 | **25.4** |
| sachet | 20.5 | 20.6 | 21.1 | **23.1** |
| straw | 26.3 | 23.1 | **27.3** | 26.2 |
| trash-bin | 55.9 | **59.2** | 52.2 | 50.6 |
| mAP$_{50}$ | **40.0** | 38.7 | 37.8 | 36.9 |
| mAP | **25.0** | 24.2 | 23.2 | 23.2 |
| mAP$_{(small)}$ | 13.5 | 12.8 | - | **14.7** |
| mAP$_{(medium)}$ | 23.2 | 23.8 | - | **27.6** |
| mAP$_{(large)}$ | **33.0** | 29.3 | - | 27.6 |

We assume misclassification of the detected instances is the reason for the low accuracy in 10-class dataset. It can be observed that mAP of 4-class set is significantly higher than that of the 10-class set, suggesting that the main bottleneck lies at the classification stage rather than detection. The major false positives observed were leaves and rocks. Some of them look especially similar to plastic litter which are coated with dust, making it difficult to differentiate through human interpretation too.

We also have evaluated our Faster R-CNN and YOLO-v5l models separately trained on PlastOPol, TACO, and pLitterStreet datasets. The resulted metrics for this evaluation are presented in Table 4. Both the models trained on PlastOPol, and TACO as single class detectors. And the same both were trained on the 4-class pLitterStreet dataset, the cross evaluation on PlastOPol and TACO was performed as a single class. Instances in the TACO dataset originally belong to multiple categories, and they are remapped to single category for the evaluation. We combined all the instances of pLitterStreet 4-class test set into a common class, except trash bin. This is to facilitate the evaluation of models trained on PlastOPol and TACO datasets, since they are single class object detectors.

Table 4 Evaluation of Faster R-CNN and YOLO-v5l models separately trained on PlastOPol, TACO and pLitterStreet(ours).

| | PlastOPol | TACO | pLitterStreet |
|---|---|---|---|
| Faster R-CNN + trained on PlastOPol | 63.8 | 36.9 | **7.6** |
| YOLO-v5l + trained on PlastOPol | 64.1 | 31.4 | **8.1** |
| Faster R-CNN + trained on TACO | 55.1 | 43.6 | **7.9** |
| YOLO-v5l + trained on TACO | 55.0 | 45.4 | **12.5** |
| Faster R-CNN + trained on pLitterStreet (4-class) | **25.9** | **24.4** | 40.9 |
| YOLO-v5l + trained on pLitterStreet (4-class) | **25.7** | **32.1** | 44.0 |

### 4.2 Result of Litter Hotspots Mapping

Our study involved mapping five cities in three different countries: Sri Lanka, Thailand, and Vietnam. We selected these locations based on their varying levels of urbanization and waste management practices. Specifically, in Thailand, we mapped one city twice, at two different periods, to examine changes in street-level litter and trash bin distributions. Meanwhile, in Sri Lanka and Vietnam, we mapped three cities one time understanding of litter accumulation patterns. Result of litter hotspots mapping is shown in Figure 6.

Our results indicate that trash bin distribution varies considerably between cities, with Can Tho city in Vietnam and Chiang Rai and Ubon Ratchathani cities in Thailand having more evenly distributed bins compared to the two cities, we surveyed in Sri Lanka. In Hanwella and Mawanella cities in Sri Lanka, litter was found to be distributed across many streets, and in some areas with high trash volumes, there were no trash bins present.

In the case of Chiang Rai where we mapped twice, we found that litter and trash bin distribution remained consistent in Chiang Rai across both time periods. In contrast, while trash bin distribution in Ubon - Ratchathani was consistent between July 2021 and January 2022, the distribution of street-side litter changed notably. Specifically, we observed a greater accumulation of litter in the northern part of the city in January 2022 compared to July 2021. This information can inform policymakers and city managers to adapt their litter collection and waste management strategies to respond to changes in litter accumulation patterns.

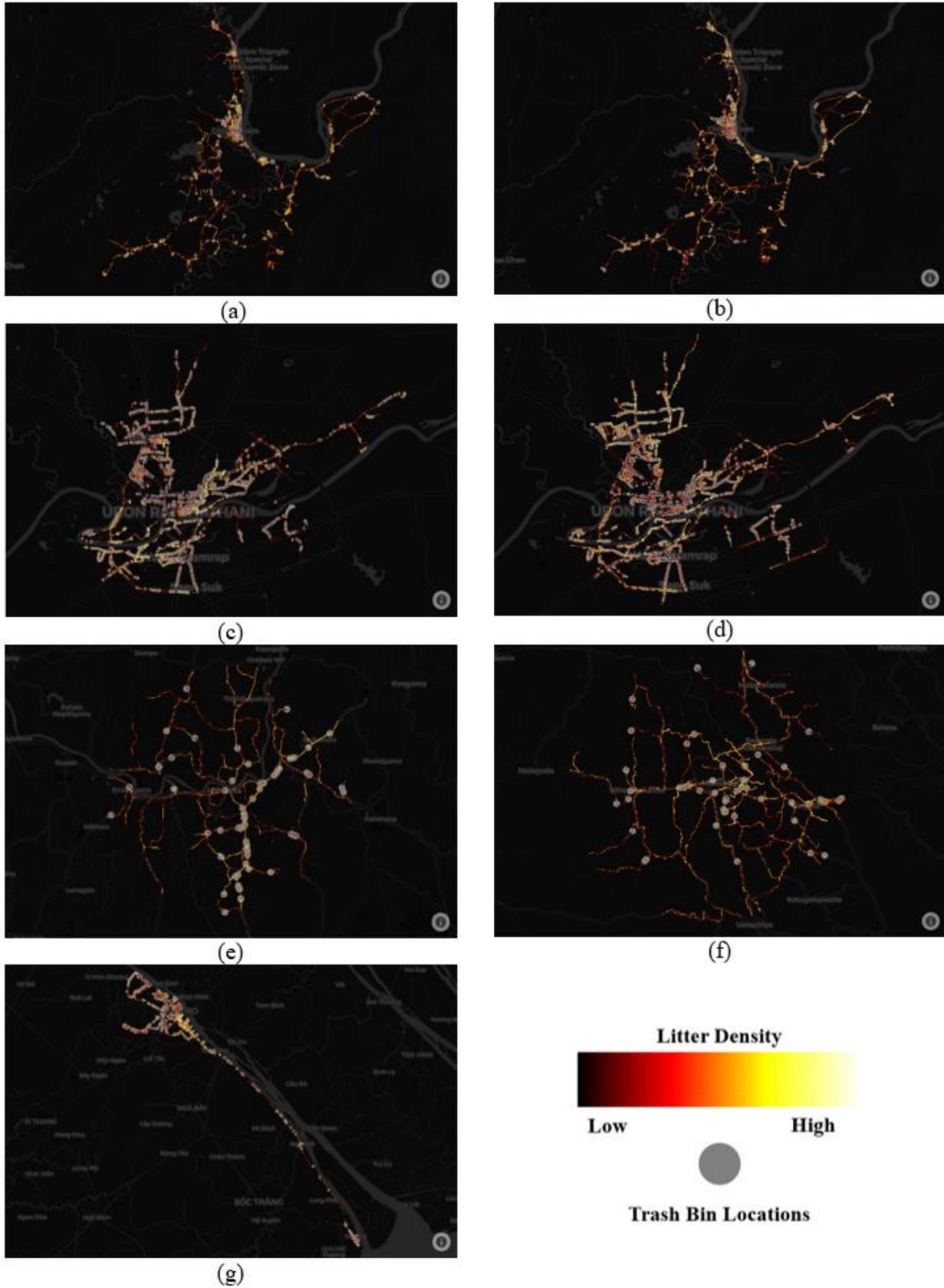

**Fig. 6** Result of litter hotspots mapping (a) Chiang Rai city, Thailand in September 2021, (b) Chiang Rai, Thailand

city in February 2022, (c) Ubon Ratchathani city, Thailand in July 2021, (d) Ubon Ratchathani city, Thailand in February 2022, (e) Hanwella city, Sri Lanka in January 2022, (f) Mawanella city, Sri Lanka in January 2022, and (g) Can Tho city, Vietnam in January 2022. (*Base Map: © Carto © OpenStreetMap contributors*)

## 5. Discussion

Plastic litter detection is a challenging task for deep learning algorithms due to the complex nature of plastic litter. The primary reason for the complexity is the extreme variation observed for typical object characteristics like shape, size, and color of plastic litter. Over time these characteristics can change due to weathering or breakdown, compounding this problem. For example, when littered, a plastic bag might be torn into pieces, or its color would be changed and contaminated or coated with dust or soil.

The nature of the plastic litter dataset is open-ended with a seemingly unlimited feature set, especially since new types of plastic objects with unobserved characteristics are created every day. Therefore, it is not realistic to sort the objects seen in the images into a list of categories. The best possible approach is categorizing plastic litter according to items commonly observed in the surroundings of the area of interest. The other challenges are similarities between plastic litter and non-plastic litter, which led to incorrect interpretation as plastic litter and caused a drop in accuracy.

Crowdsourced datasets such as TACO and PlastOPol mostly consist of images captured with handheld cameras such as mobile phones, where the users have the flexibility to move closer to the litter objects, resulting in high quality images. However, this handheld camera approach is labour intensive and may not be feasible for conducting city scale mapping of plastic pollution. In our approach, we utilized vehicle mounted cameras for data collection to enable the practical mapping of plastic pollution at city scale, this approach introduced more challenges. Street level litter appears smaller in the images captured with vehicle mounted cameras. pLitterStreet dataset is highly dominated with small size plastic litter instances which makes it much more difficult in detection and classification.

From the cross evaluation, it can be served that although the models trained on their respective train sets achieved the best results on corresponding test sets. The models trained on our dataset achieved better accuracy on PlastOPol and TACO test sets than the accuracy yielded on our test set with models trained on PlastOPol or TACO. This indicates that models trained on our dataset is more generalized than others and emphasizes the importance of this dataset in street level plastic litter detection.

The pLitterStreet dataset currently includes images only from Sri Lanka, Thailand, and Vietnam. City-scale plastic littering heatmaps were prepared using the detections from trained plastic litter detectors to understand scalability of this dataset. The produced heatmaps appear to be reliable in representing the actual conditions of plastic pollution on the ground. However, further improvements in the detection accuracy could proportionally enhance the quality of these heatmaps. Chiang Rai and Ubon Ratchathani cities in Thailand were mapped twice, and it was observed that results are consistent and reliable. Moreover, conducting recitative mapping of cities helps in identifying temporal changes in plastic littering hotspots.

Overall, our findings highlight the need for improved waste management practices in urban areas, particularly in areas with high litter accumulation. By mapping litter hotspots and trash bin distributions, our methodology provides a valuable tool for policymakers and city planners to identify areas in need of targeted interventions to address this pressing environmental issue.

## 6. Conclusion

This paper presents a novel methodology for mapping street-level plastic litter hotspots, which was developed and implemented using our own open-source dataset. Our methodology consists of four main steps: data collection, image annotation, model training and heatmap generation. The dataset we used in this study was collected using a vehicle-mounted camera equipped with GPS location tracking capabilities, which enabled us to capture geospatial information for each recorded frame. We then utilized a team of annotators and open-source annotation tools to label a subset of the collected data, which was used to train a set of deep learning algorithms. The trained models were then used to predict the presence of litter in new video frames, and these predictions were combined with corresponding GPS locations to produce heat maps of cities, highlighting areas with high litter accumulation.

Additionally in this paper, we also introduced pLitterStreet, the dataset for street level plastic litter detection. The pLitterStreet dataset consists of more than 13,000 images collected from vehicle-mounted cameras. pLitterStreet consists of nearly 80,000 labels categorized into 10 classes. We evaluated the performance of some of the popular object detection algorithms on our dataset. Results indicated average accuracy metrics to detect plastic, piles, face masks and trash bins, which were adequate to meet the practical application of generating a city's plastic littering snapshot successfully. We hope that by publishing the dataset, it could be expanded with data from new cities and to generalize the dataset for global application. Overall, our methodology and dataset offer a new approach for mapping plastic litter hotspots that can be utilized by researchers and practitioners alike to improve our understanding of the extent and distribution of plastic waste in urban environments.

## 7. Future work

We assume that the novelty of plastic items may cause any dataset in this domain to be outdated as new and different looking plastic items are created every day. Therefore, the dataset should be updated with new samples to reflect new types of plastic litter. To improve the accuracy of deep learning algorithms for litter detection, it is necessary to prepare a more comprehensive and diverse dataset that includes various types of litter and different environments. It is recommended that more locations should be included to generalize the dataset to detect the plastics litter in the images from across the globe. As the dataset size grows, adding more categories to the training dataset is a top priority. It could help understanding the composition of plastic items which are dominant in littering.

Future studies can also explore pooling and using data from different sources and cameras, including mobile cameras, to increase the quantity and diversity of data for litter detection. Moreover, a key area of future research could be focused on developing light-weight models that can perform with high accuracy on resource-constrained devices. Finally, efforts can be made to reduce false positives by incorporating techniques that can differentiate litter from similar objects commonly found in street sides and roads, such as organic matter and other debris.

**References**

1. Ren, S., He, K., Girshick, R., & Sun, J. (2015). Faster R-CNN: Towards real-time object detection with region proposal networks. In *Advances in neural information processing systems*.
2. Redmon, J., & Farhadi, A. (2018). Yolov3: An incremental improvement. *arXiv preprint arXiv:1804.02767*.
3. Lin, T. Y., Goyal, P., Girshick, R., He, K., & Dollár, P. (2017). Focal loss for dense object detection. In *Proceedings of the IEEE international conference on computer vision* (pp. 2980-2988).
4. Lynch, S. (2018). OpenLitterMap.com – open data on plastic pollution with blockchain rewards (littercoin). *Open Geospatial Data, Software and Standards*, *3*(1), 1-10.
5. Rad, M. S., von Kaenel, A., Droux, A., Tieche, F., Ouerhani, N., Ekenel, H. K., & Thiran, J. P. (2017). A computer vision system to localize and classify wastes on the streets. In *Computer Vision Systems: 11th International Conference, ICVS 2017, Shenzhen, China, July 10-13, 2017, Revised Selected Papers 11* (pp. 195-204). Springer International Publishing.
6. Lin, T. Y., Maire, M., Belongie, S., Hays, J., Perona, P., Ramanan, D., ... & Zitnick, C. L. (2014). Microsoft coco: Common objects in context. In *Computer Vision–ECCV 2014: 13th European Conference, Zurich, Switzerland, September 6-12, 2014, Proceedings, Part V 13* (pp. 740-755). Springer International Publishing.
7. Everingham, M., Van Gool, L., Williams, C. K., Winn, J., & Zisserman, A. (2010). The pascal visual object classes (voc) challenge. *International journal of computer vision*, *88*, 303-338.
8. Cordts, M., Omran, M., Ramos, S., Rehfeld, T., Enzweiler, M., Benenson, R., … & Schiele, B. (2016). The cityscapes dataset for semantic urban scene understanding. In *Proceedings of the IEEE conference on computer vision and pattern recognition* (pp. 3213-3223).
9. Neuhold, G., Ollmann, T., Rota Bulo, S., & Kontschieder, P. (2017). The mapillary vistas dataset for semantic understanding of street scenes. In *Proceedings of the IEEE international conference on computer vision* (pp. 4990-4999).
10. Geiger, A., Lenz, P., Stiller, C., & Urtasun, R. (2013). Vision meets robotics: The kitti dataset. *The International Journal of Robotics Research*, *32*(11), 1231-1237.
11. Proença, P. F., & Simoes, P. (2020). Taco: Trash annotations in context for litter detection. *arXiv preprint arXiv:2003.06975*.

12. Córdova, M., Pinto, A., Hellevik, C. C., Alaliyat, S. A. A., Hameed, I. A., Pedrini, H., & Torres, R. D. S. (2022). Litter detection with deep learning: A comparative study. *Sensors*, *22*(2), 548.
13. Wang, J., Guo, W., Pan, T., Yu, H., Duan, L., & Yang, W. (2018, July). Bottle detection in the wild using low-altitude unmanned aerial vehicles. In *2018 21st International Conference on Information Fusion (FUSION)* (pp. 439-444). IEEE.
14. Kraft, M., Piechocki, M., Ptak, B., & Walas, K. (2021). Autonomous, onboard vision-based trash and litter detection in low altitude aerial images collected by an unmanned aerial vehicle. *Remote Sensing*, *13*(5), 965.
15. Lynch, S. (2018). OpenLitterMap. com–open data on plastic pollution with blockchain rewards (littercoin). *Open Geospatial Data, Software and Standards*, *3*(1), 1-10.
16. Majchrowska, S., Mikołajczyk, A., Ferlin, M., Klawikowska, Z., Plantykow, M. A., Kwasigroch, A., & Majek, K. (2022). Deep learning-based waste detection in natural and urban environments. *Waste Management*, *138*, 274-284.
17. Brooks, J. (2019). COCO Annotator. https://github.com/jsbroks/coco-annotator/. https://github.com/jsbroks/coco-annotator/
18. Jocher, G., Stoken, A., Borovec, J., NanoCode012, ChristopherSTAN, Changyu, L., Laughing, tkianai, Hogan, A., lorenzomammana, yxNONG, AlexWang1900, Diaconu, L., Marc, wanghaoyang0106, ml5ah, Doug, Ingham, F., Frederik, … Rai, P. (2020). ultralytics/yolov5: v3.1 - Bug Fixes and Performance Improvements (v3.1) [Computer software]. Zenodo. https://doi.org/10.5281/zenodo.4154370
19. Wu, Y., Kirillov, A., Massa, F., Lo, W.-Y., & Girshick, R. (2019). Detectron2. https://github.com/facebookresearch/detectron2. https://github.com/facebookresearch/detectron2
20. Jocher, G., Kwon, Y., guigarfr, perry0418, Veitch-Michaelis, J., Ttayu, Suess, D., Baltacı, F., Bianconi, G., IlyaOvodov, Marc, e96031413, Lee, C., Kendall, D., Falak, Reveriano, F., FuLin, GoogleWiki, Nataprawira, J., … Xinyu, W. (2020). ultralytics/yolov3: v8 - Final Darknet Compatible Release (Version v8) [Computer software]. Zenodo. https://doi.org/10.5281/zenodo.4279923
21. Paszke, A., Gross, S., Massa, F., Lerer, A., Bradbury, J., Chanan, G., Killeen, T., Lin, Z., Gimelshein, N., Antiga, L., Desmaison, A., Kopf, A., Yang, E., DeVito, Z., Raison, M., Tejani, A., Chilamkurthy, S., Steiner, B., Fang, L., … Chintala, S. (2019). PyTorch: An Imperative Style, High-Performance Deep Learning Library. In *Advances in Neural Information Processing Systems 32* (pp. 8024–8035).